%% file: main.tex
\documentclass[10pt,twocolumn,letterpaper]{article}

\usepackage{makecell}
\usepackage[accsupp]{axessibility}

\usepackage{cvpr}              
\usepackage{tikz}
\usepackage{pgfplots}
\usepackage{colortbl}
\usetikzlibrary{positioning, shapes.geometric, arrows, patterns, decorations.pathreplacing, calc}
\usetikzlibrary{positioning, shapes.geometric, arrows, patterns, decorations.pathreplacing, calc}
\pgfplotsset{compat=1.17}

\definecolor{cvprblue}{rgb}{0.21,0.49,0.74}
\definecolor{ballblue}{rgb}{0.13, 0.67, 0.8}
\definecolor{bleudefrance}{rgb}{0.19, 0.55, 0.91}
\usepackage[pagebackref,breaklinks,colorlinks,allcolors=cvprblue]{hyperref}

\def\paperID{18568} 
\def\confName{CVPR}
\def\confYear{2025}

\title{Toward Real-world BEV Perception: Depth Uncertainty Estimation \\ via Gaussian Splatting}

\author{
Shu-Wei Lu$^{1}$ \quad Yi-Hsuan Tsai$^{2}$ \quad Yi-Ting Chen$^{1}$\thanks{: Corresponding author.} \\
$^{1}$National Yang Ming Chiao Tung University \quad $^{2}$Atmanity Inc.
}

\begin{document}
\maketitle
\input{sec/0_abstract}  
\input{sec/1_intro}
\input{sec/2_related_work}
\input{sec/FIG3}
\input{sec/3_method}
\input{sec/4_experiment}
\input{sec/5_conclusion}
{
    \paragraph{Acknowledgements.}The work is sponsored in part by the Higher Education Sprout Project of the National Yang Ming Chiao Tung University and Ministry of Education (MOE), the Yushan Fellow Program Administrative Support Grant, and the National Science and Technology Council (NSTC) under grants 111-2218-E-A49-029-, 112-2634-F-002-006-, and Institute for Information Industry (III).
    \small
    \bibliographystyle{ieeenat_fullname}
    \bibliography{main.bbl}
}
\clearpage
\input{X_suppl}
\end{document}

%% file: sec/0_abstract.tex
\begin{abstract}
Bird's-eye view (BEV) perception has gained significant attention because it provides a unified representation to fuse multiple view images and enables a wide range of downstream autonomous driving tasks, such as forecasting and planning.
%
%
%
Recent state-of-the-art models utilize projection-based methods which formulate BEV perception as query learning to bypass explicit depth estimation.
%
%
While we observe promising advancements in this paradigm, they still fall short of real-world applications because of the lack of uncertainty modeling and expensive computational requirement.
%
%
In this work, we introduce GaussianLSS, an uncertainty-aware BEV perception framework that revisits the unprojection-based method, specifically the Lift-Splat-Shoot (LSS) paradigm, and enhances it with depth uncertainty modeling. Our GaussianLSS represents spatial dispersion by learning a soft depth mean and computing the variance of the depth distribution, which implicitly captures object extents. We then transform the depth distribution into 3D Gaussians and rasterize them to construct uncertainty-aware BEV features.
We evaluate GaussianLSS on the nuScenes dataset, achieving state-of-the-art performance compared to unprojection-based methods.
In particular, it provides significant advantages in speed, running 2.5x faster, and in memory efficiency, using 0.3x less memory compared to projection-based methods, while achieving competitive performance with only a 0.4\% IoU difference. See our project page for more details: \href{https://hcis-lab.github.io/GaussianLSS/}{https://hcis-lab.github.io/GaussianLSS/}.
\end{abstract}

%% file: sec/1_intro.tex
%
\section{Introduction}
\label{sec:intro}
\input{sec/FIG1}
Bird’s-eye view (BEV) perception is an emerging and crucial technique in autonomous driving, offering a unified spatial representation for integrating multiple sensor inputs. It serves as a foundation for 3D perception tasks such as 3D object detection \cite{li2022bevformer, liu2022petrv2, xie2022m} and BEV segmentation \cite{philion2020lift, fiery2021, CVT, harley2022simple, chambon2024pointbev}, which are essential for understanding the driving environment. BEV perception also plays a pivotal role in downstream applications, including motion forecasting \cite{vip3d} and planning \cite{hu2023_uniad, fiery2021, hu2022stp3}, where accurate spatial understanding is critical for safety and decision-making. In addition, BEV facilitates effective multi-modality fusion by providing a robust intermediate representation for sensor integration \cite{liu2022bevfusion, harley2022simple}.

Existing approaches for BEV perception can be broadly categorized into two paradigms: (1) 2D unprojection methods, which estimate depth and unproject features into 3D space \cite{philion2020lift, fiery2021, liu2022bevfusion}, and (2) 3D projection methods, which project predefined 3D coordinate volumes onto camera views and aggregate image features \cite{li2022bevformer, harley2022simple, chambon2024pointbev}. While these paradigms have driven significant progress, they often involve trade-offs in accuracy, computational cost, and scalability, limiting their applicability in real-world scenarios. 3D projection approaches represent the state-of-the-art in terms of accuracy; however, their reliance on 3D grids leads to high computational costs, making them less practical for real-time applications. 
%
To overcome these limitations, we introduce \textit{GaussianLSS}, a 2D unprojection-based approach that balances accuracy and efficiency to meet the real-time requirements of autonomous driving applications.

The proposed GaussianLSS addresses the fundamental challenge of depth estimation by introducing a novel depth uncertainty modeling technique. We begin with probabilistic depth estimation, calculating the variance within the depth distribution to capture uncertainty. This variance indicates the degree of spread around the mean depth, reflecting the level of confidence in depth estimation at each depth bin. By modeling depth uncertainty in this way, precise depth estimates become less critical, as the extent of the spread naturally encodes object boundaries and extents, improving the robustness of BEV representations even when depth is uncertain.
Subsequently, we transform the depth distribution into a 3D probability distribution by unprojecting each pixel’s coordinates across the depth bins. This process seamlessly maps depth uncertainty into 3D space, where we utilize Gaussian Splatting rasterization for efficient BEV feature aggregation. This approach enables smooth, uncertainty-aware rendering of features across the BEV plane, capturing object extents and spatial relationships with high fidelity. The performance of GaussianLSS is shown in Figure~\ref{fig:comparison_depth_grid}.

The main contributions of our work are as follows:
\begin{itemize}
    \item We introduce GaussianLSS, a novel depth uncertainty modeling approach tailored for BEV perception, which captures and leverages depth ambiguity for improved spatial representation.

    \item We propose a computationally efficient method to transform depth uncertainty into a 3D probability distribution, seamlessly integrating it with Gaussian Splatting for fast and accurate BEV feature aggregation.

    \item Our GaussianLSS achieves state-of-the-art results among 2D unprojection approaches and is competitive with 3D projection methods. Moreover, it significantly reduces memory usage and inference time, making it well-suited for real-world autonomous driving applications.
\end{itemize}

%% file: sec/FIG1.tex
\begin{figure}[t]
    \centering
    \includegraphics[width=1\linewidth]{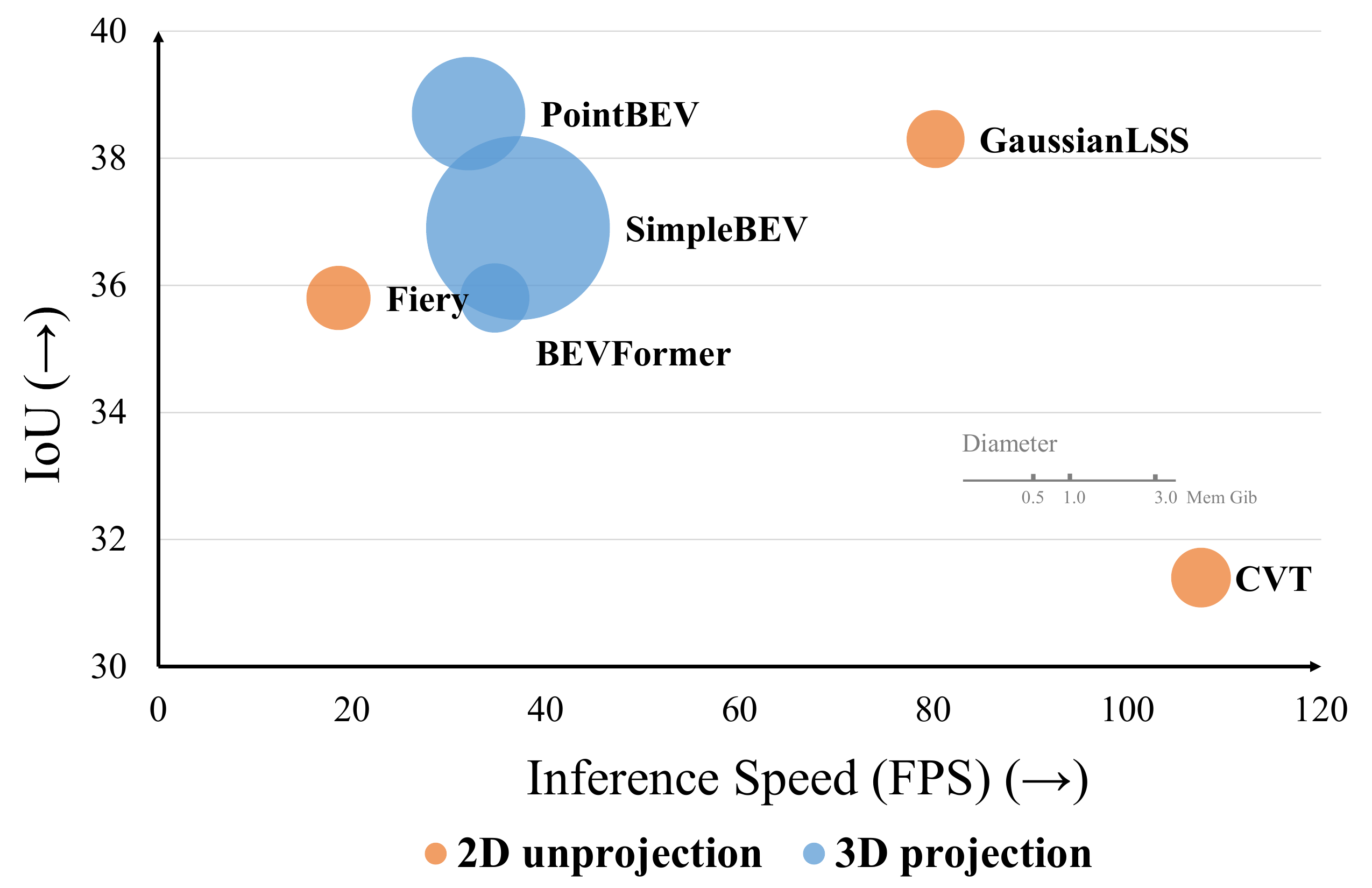}
    \caption{\textbf{Comparisons of 2D unprojection-based and 3D projection-based methods on vehicle BEV segmentation.} GaussianLSS achieves state-of-the-art performance among 2D unprojection baselines. In addition, it also demonstrates competitive performance compared to 3D projection-based methods, while offering significant advantages in memory efficiency and inference speed.}
    \label{fig:comparison_depth_grid}
\end{figure}

%% file: sec/2_related_work.tex
\section{Related Work} 
\label{sec:Related} 
Since 3D object detection and BEV segmentation share considerable overlap, we include relevant 3D object detection approaches in our discussion of related works.

\vspace{-2mm}
\paragraph{3D projection.} 3D projection methods map predefined 3D voxel points onto the image plane to sample features, eliminating the need for explicit depth estimation. This approach bypasses the complexities of direct depth prediction by placing features at plausible 3D locations. Notable methods, such as BEVFormer \cite{li2022bevformer} and SimpleBEV \cite{harley2022simple}, employ grid sampling to efficiently aggregate multi-view features across the BEV plane. To address the challenges associated with grid resolution, PointBEV \cite{chambon2024pointbev} introduces a coarse-to-fine training strategy, transitioning from dense to sparse grids, which reduces memory consumption while preserving accuracy.

Despite these advancements, 3D projection methods remain computationally intensive compared to 2D unprojection approaches, limiting their scalability in real-world applications.

\vspace{-2mm}
\paragraph{Implicit 2D unprojection.} Implicit 2D unprojection methods \cite{CVT, liu2022petr, liu2022petrv2, Pan_2023_CVPR, bartoccioni2022lara} leverage transformer-based architectures and MLPs to learn the mapping from 2D images to 3D space without explicitly predicting depth. These approaches focus on implicitly learning spatial relationships and depth cues through the integration of BEV grid-like queries and camera-aware positional embeddings in a cross-attention framework.

However, without explicit depth estimation, these methods encounter challenges with depth ambiguity, as the projection is only implicit. Furthermore, their computational complexity increases significantly with larger BEV grid and image resolutions, making them less efficient and scalable for high-resolution applications. These limitations restrict their practicality for detailed scenes requiring precise spatial representations.

\vspace{-2mm}
\paragraph{Explicit 2D unprojection.} Lift-Splat-Shoot (LSS) \cite{philion2020lift} introduces an efficient pipeline to lift 2D features into 3D, a design that has become the foundation in 3D perception tasks \cite{liu2022bevfusion, xie2022m, fiery2021, hu2022stp3}. This approach relies heavily on accurate depth estimation, which makes it sensitive to depth prediction errors that can propagate into the BEV representation. To mitigate this, subsequent works have added depth supervision as an auxiliary loss to improve depth accuracy \cite{li2022bevdepth, li2022bevstereo, li2023fbbevbevrepresentationforwardbackward, li2024bevnextrevivingdensebev}.

Although these methods use probabilistic depth distributions to softly lift features, they typically lack an explicit representation of depth uncertainty. This limitation hinders their ability to handle depth ambiguities effectively, particularly in complex scenarios. GaussianLSS addresses this by explicitly modeling depth uncertainty as the variance of the probabilistic distribution. This uncertainty-aware depth representation reduces the reliance on precise depth estimation by allowing the model to capture varying spatial extents spread around the depth mean, as shown in Figure~\ref{fig:DepthUncertainty}.

\input{sec/FIG2}

\vspace{-2mm}
\paragraph{Uncertainty modeling.} Uncertainty modeling is a widely adopted approach to capture ambiguity in computer vision tasks, with applications in areas such as semantic segmentation \cite{huang2018efficientuncertaintyestimationsemantic, Mukhoti_2023_CVPR}, monocular depth estimation \cite{Poggi_CVPR_2020, Hornauer2022GradientbasedUF, shao2023IEBins, hu2022uncertainty}, and novel-view synthesis \cite{xue2024neuralvisibilityfielduncertaintydriven, Ren2024NeRF, shen2023estimating3duncertaintyfield}.
Common methods for estimating uncertainty include: \begin{itemize} \item Variance of Predicted Distributions: Measure uncertainty based on the variance of predicted probability distributions, providing a direct indication of confidence in the output \cite{shao2023IEBins, hu2022uncertainty}. \item MLP-based Uncertainty Estimation: Use a multi-layer perceptron (MLP) to output a single uncertainty score or predict a distribution characterized by mean and variance, where the variance serves as the uncertainty measure \cite{Ren2024NeRF, liu2024difflow3d, GuSongEtAl2024}. \item Bayesian Networks: Incorporate probabilistic prior distributions to model uncertainty in a principled framework \cite{xue2024neuralvisibilityfielduncertaintydriven, lu2024quantifyinguncertaintymotionprediction}. \end{itemize}

Each of these methods provide unique ways to model uncertainty, supporting more robust predictions across varied scenarios. In our work, we adopt uncertainty modeling by focusing on depth distribution variance, leveraging it to enhance BEV segmentation performance, particularly in the presence of depth ambiguity.

%% file: sec/FIG2.tex
\begin{figure*}[t]
    \centering
    \includegraphics[width=\linewidth]{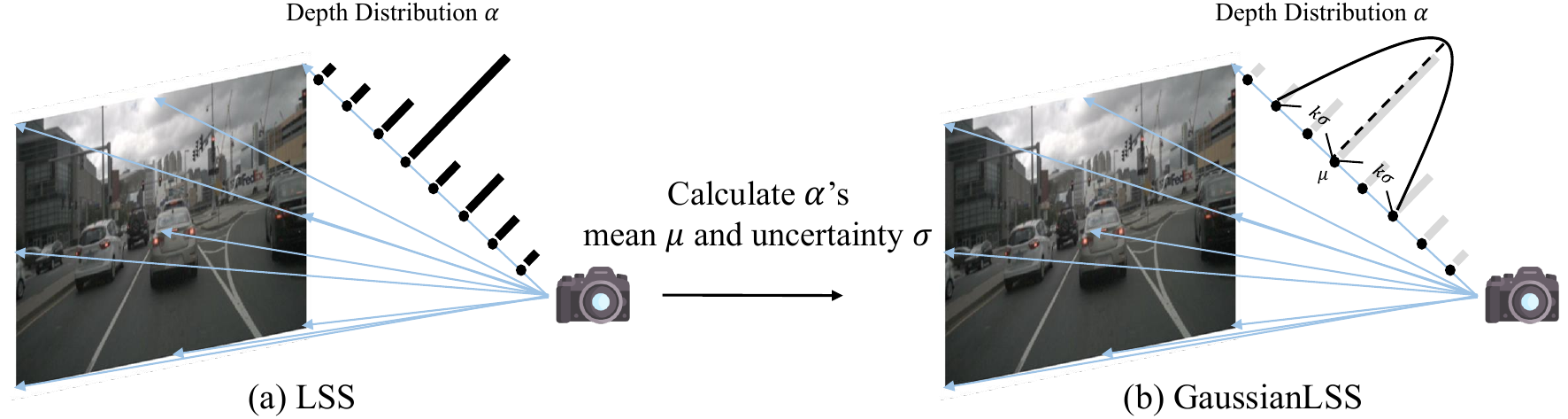}
    \caption{\textbf{Comparison between the lifting method of Lift-Splat-Shoot (LSS) \cite{philion2020lift} and our proposed GaussianLSS.} LSS uses discrete depth probabilities for soft depth weighting but struggles with depth ambiguity due to the inherently ill-posed nature of depth estimation. GaussianLSS addresses depth ambiguity by modeling depth uncertainty. We calculate the depth mean ($\mu$) and uncertainty ($\sigma$) of the predicted depth distribution, converting the original soft weighting to an uncertainty-aware range $[\mu-k\sigma,\mu+k\sigma]$. The parameter $k$ acts as an error tolerance coefficient to control the extent of the spread centered at the mean depth.}
    \label{fig:DepthUncertainty}
    \vspace{3mm}
\end{figure*}

%% file: sec/FIG3.tex
\begin{figure*}[ht]
    \centering
    \includegraphics[width=0.9\linewidth]{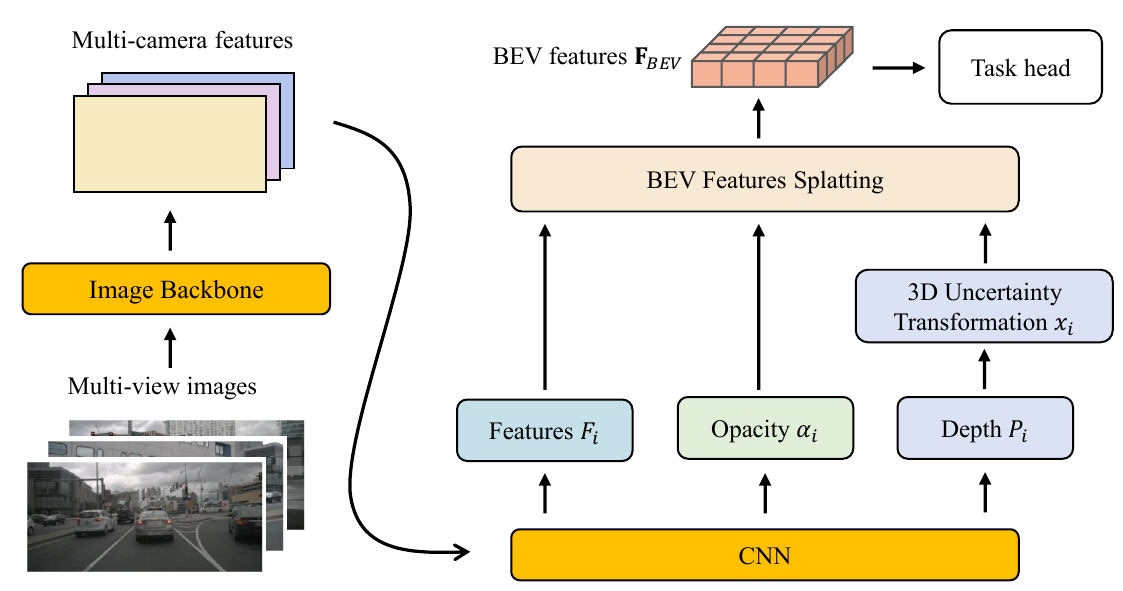}
    \caption{\textbf{Overview of GaussianLSS.} Multi-view images are first processed by a backbone network to extract features. They are then input to a simple CNN layer to obtain splat features $F_i$, opacity $\alpha_i$, and depth distribution $P_i$. The predicted depth distribution undergoes an uncertainty transformation to produce a 3D uncertainty $x_i$. Next, BEV features are obtained through a splatting process, integrating features across views. The resulting BEV features $\mathbf{F}_{\text{BEV}}$, enriched with uncertainty awareness, are used as input to the task-specific head for prediction.}
    \label{fig:architecture}
    \vspace{3mm}
\end{figure*}

%% file: sec/3_method.tex
\section{GaussianLSS}
\label{sec:Method}
We aim to address the challenge of depth ambiguity in real-world scenarios by incorporating depth uncertainty modeling into the BEV representation pipeline. The overview of GaussianLSS is shown in Figure~\ref{fig:architecture}.

GaussianLSS begins by predicting a per-pixel depth distribution, providing an estimate of the corresponding depth uncertainty (see Section~\ref{sec:3.1}). Using the camera's projection matrix, we define a camera frustum space, transforming this depth uncertainty into a 3D distribution represented as a Gaussian with mean and covariance matrix (see Section~\ref{sec:3.2}). To achieve efficient BEV feature splatting, we introduce an opacity parameter into the 3D Gaussian representation, enabling the use of Gaussian Splatting rasterization. However, we observe that BEV features can be distorted due to inconsistent depth means between adjacent pixels. To address this issue, we adopt a multi-scale BEV rendering approach (see Section~\ref{sec:3.3}).

\subsection{Depth Uncertainty Modeling}
\label{sec:3.1}
We first revisit the pioneering work Lift-Splat-Shoot\cite{philion2020lift}. It first discretizes the depth range \([d_{\text{min}}, d_{\text{max}}]\) into \( B \) bins. This creates a set of discrete depths $D$:
\[
D = \left\{ d_i = d_{\text{min}} + i \cdot \frac{d_{\text{max}} - d_{\text{min}}}{B} \right\}_{i=0}^{B-1}.
\]
Then we associate $D$ with pixel coordinates to create a camera frustum space $\mathcal{C} \in \mathbb{R}^{H \times W \times B \times 3}$. For each pixel \( p \) in the image, the network predicts a context vector \( \mathbf{c} \in \mathbb{R}^C \) and a depth distribution \( \boldsymbol{\alpha} \in \Delta^{B} \), where \(\Delta^{B}\) represents the \( B \)-dimensional probability simplex. For a given depth \( d \), the feature \( \mathbf{c}_d \in \mathbb{R}^C \) associated with point \( p_d \) in the frustum space is defined as the context vector scaled by the corresponding depth distribution coefficient \( \alpha_d \):
\[
\mathbf{c}_d = \alpha_d \mathbf{c}.
\]
However, this soft assignment mechanism has several disadvantages:
\begin{itemize}
    \item Sparse BEV projection: The discontinuity in discretized depths generates a sparse BEV projection, leading to incomplete spatial coverage in the BEV representation.

    \item Unstable depth distribution: The depth distribution is obtained via a softmax operation, but softmax can yield vastly different probabilities for nearby depth bins \cite{https://doi.org/10.48550/arxiv.2302.12288}. This results in inconsistent BEV features, as similar depths may receive disproportionate attention due to slight changes in depth values.
\end{itemize}
Hence, we propose an uncertainty-aware depth modeling approach that mitigates these issues by introducing a continuous depth representation and explicitly modeling depth uncertainty, allowing for smoother and more reliable BEV feature aggregation.

Inspired by \cite{shao2023IEBins}, we capture depth uncertainty by modeling the variance of the depth distribution. Let $P$ be the per-pixel depth distribution, for each pixel $p$, we calculate the depth mean $\mu = \sum_{i=0}^{B-1}P_i(p) d_i$ and variance $\sigma^2 = \sum_{i=0}^{B-1}P_i(p) (d_i-\mu)^2$ 
, where $d_i$ and $P_i(p)$ denote the depth value of $i$-th bin and its probability. Using an error tolerance coefficient $k$, we define a soft depth estimation range $\mathbf{\hat{D}}=[\mu-k\sigma, \mu+k\sigma]$. This range accommodates the depth uncertainty, allowing for more flexible and reliable depth projection by capturing the distribution’s spread around the mean (see Figure~\ref{fig:DepthUncertainty}).

\subsection{3D Uncertainty Transformation}
\label{sec:3.2}
We generate a soft depth estimation \( \mathbf{\hat{D}} \) in the depth space, which we then transform into a 3D representation. Given a point \( p = (u, v, d) \) in the frustum space \( \mathcal{C} \), where \( u \) and \( v \) represent the pixel coordinates and \( d \) is the depth, we unproject \( p \) into 3D coordinates using the camera intrinsics \( I \) and extrinsics \( E \) as follows:
\[
p^{3d} = E^{-1} \cdot \left( d \cdot I^{-1} \cdot[u,v,1]^T  \right),
\]
where \( p^{3d} \in \mathbb{R}^3 \) is the unprojected 3D point in the ego-vehicle frame.
Then we calculate the 3D mean \( \mu_{3d} \) and covariance \( \Sigma \) for point \( p \) by:

\begin{equation}
\label{equa:3}
    \mu_{3d} = \sum_{i=0}^{B-1} P_i(p) \, p^{3d}_i,
\end{equation}

\begin{equation}
\label{equa:4}
    \Sigma = \sum_{i=0}^{B-1} P_i(p) \, (p^{3d}_i - \mu_{3d})(p^{3d}_i - \mu_{3d})^T,
\end{equation}
where \( P_i(p) \) is the probability associated with each depth bin \( i \), and \( p^{3d}_i \) denotes the 3D point corresponding to depth \( d_i \) for pixel \( p \). Thus, the soft depth estimation \( \mathbf{\hat{D}} \) is transformed into a 3D Gaussian, with the mean \( \mu_{3d} \) representing the center of the distribution in 3D space, and \( \Sigma \) describing its spread. As described in Section \ref{sec:3.1}, we apply an error tolerance coefficient \( k \) to define a 3D uncertainty range around \( \mu_{3d} \). This soft 3D range is represented by an ellipsoid centered at \( \mu_{3d} \), defined by:

\[
(\mathbf{x} - \mu_{3d})^T \Sigma^{-1} (\mathbf{x} - \mu_{3d}) \leq k^2,
\]
where \( \mathbf{x} \) is any point in the 3D space. This ellipsoid captures the uncertainty in the 3D position of \( p \) based on the depth estimation's spread. We denote this transformation $T$ as:
\[
x_i=T(\mathbf{\hat{D}},I,E)=(\mu_{3d},\Sigma),
\]
where $x_i$ represents the transformed 3D uncertainty.

\subsection{BEV Features Splatting}
\label{sec:3.3}
In this section, we describe how to efficiently splat depth uncertainty, integrated with predicted features, into the BEV representation.

\paragraph{Gaussian Splatting.}
Gaussian Splatting, as introduced in \cite{kerbl3Dgaussians}, models 3D scenes using 3D Gaussians $G$ defined by a 3D mean \( \mu \in \mathbb{R}^{3} \), a 3D covariance matrix \( \Sigma \in \mathbb{R}^{3 \times 3} \), and an opacity \( \alpha \in \mathbb{R}^{+} \):
\begin{equation}
    G(\mathbf{x}) = \alpha \exp \left( -\frac{1}{2} (\mathbf{x} - \mu)^T \Sigma^{-1} (\mathbf{x} - \mu) \right).
\end{equation}
The Gaussians are then projected and rendered onto a 2D plane using alpha-blending:
\begin{equation}
\label{equa:blending}
\mathbf{C} = \sum_{i \in \mathcal{N}} c_i \alpha_i \prod_{j=1}^{i-1} (1 - \alpha_j),
\end{equation}
where \( c_i \) is the color of each point, and \( \mathbf{C} \) is the computed color after blending.

\paragraph{BEV feature rendering.}
For each input image from $n$ multi-view cameras, we extract multi-camera features and using a simple CNN layer to get splatting features \( F \in \mathbb{R}^{C \times H \times W} \), depth distribution \( P \in \mathbb{R}^{B \times H \times W} \), and opacities \( \alpha\in\mathbb{R}^{1 \times H \times W} \). Then, as detailed in Section~\ref{sec:3.2}, we generate per-pixel 3D representations \( X = \{x_i\}_{i\in\mathcal{C}} \), where each \( x_i \) includes 3D spatial coordinates and a covariance matrix. By integrating these 3D representations with the feature map \( F \) and opacities \( \alpha \), we obtain a set of Gaussians \( \mathcal{G} = \{g_i = (x_i, F_i, \alpha_i)\} \in \mathbb{R}^{(C+13) \times H \times W} \). Next, $n$ sets of Gaussians $\mathcal{G}$ are combined and projected to BEV plane \( \mathcal{G}_{\text{BEV}}=\{\mathcal{G}_i\}_{i=1}^n \).

The projection onto the BEV plane is implemented by slightly modifying the original projection method which we will detail in the supplementary material. Finally, we splat the features \( F \) onto the BEV plane by replacing \( c_i \) with \( F_i \) in Eq.~\eqref{equa:blending}, resulting in the following BEV feature:

\begin{equation}
    \mathbf{F}_{\text{BEV}}(\mathbf{x}) = \sum_{i \in \mathcal{G_{\text{BEV}}}} F_i \alpha_i \exp \left( -\frac{1}{2} (\mathbf{x} - \mu_i)^{\top} \Sigma_i^{-1} (\mathbf{x} - \mu_i) \right),
\end{equation}
where \( \mathbf{F}_{\text{BEV}}(\mathbf{x}) \) denotes the rendered BEV feature at each location \( \mathbf{x} \in \mathbb{R}^2 \). To address inconsistencies in depth estimation between adjacent pixels, we propose a multi-scale BEV feature rendering approach. This method projects $\mathcal{G}$ onto BEV planes at varying resolutions (e.g., $50\times50$ or $200\times200$), effectively capturing hierarchical spatial representations. The resulting multi-scale features are then upsampled and fused to match the target BEV resolution. Finally, the fused features are processed by a segmentation head to produce the final predictions.

%% file: sec/4_experiment.tex
\section{Experimental Results}

We evaluate GaussianLSS on the nuScenes dataset~\cite{nuscenes2019}, a large-scale dataset for autonomous driving that provides synchronized data from multiple sensors. It contains a total of 1000 scenarios, splitted into 750 train sets, 150 validation sets, and 150 test sets.

\subsection{Dataset and Pre-processing}

The nuScenes dataset comprises 1,000 scenes, each 20 seconds long, captured under diverse weather conditions and times of day. Each scene includes data from six cameras covering a full 360-degree field of view around the ego-vehicle. For our experiments, we use the official training and validation splits provided by nuScenes.

To ensure fair comparisons with the prior work, we include visibility filtering experiments by selecting objects with at least 40\% coverage. This filtering focuses on the evaluation for adequately visible objects. The BEV representation is defined over a grid of size \(200 \times 200\), corresponding to an area of \([-50\,\text{m}, 50\,\text{m}]\) along both the X (forward) and Y (sideways) axes relative to the ego-vehicle. Each grid cell represents a \(0.5\,\text{m} \times 0.5\,\text{m}\) area.

\subsection{Implementation Details}

GaussianLSS is trained using a combination of segmentation, centerness, and offset loss functions, specifically focal loss~\cite{lin2018focallossdenseobject}, L1 loss, and L2 loss, with respective weights \( \lambda_1 = 1 \), \( \lambda_2 = 2 \), and \( \lambda_3 = 0.1 \). We use AdamW optimizer~\cite{loshchilov2019decoupledweightdecayregularization} with a learning rate of \( 3 \times 10^{-4} \), weight decay of \( 1 \times 10^{-7} \), and a cosine learning rate scheduler. The total batch size is set to 8, distributed across 2 NVIDIA RTX 4090 GPUs, and we train GaussianLSS for 50 epochs.

The input images are resized to \( 224 \times 480 \) and \( 448 \times 800 \). For the uncertainty transformation, we utilize 1/8 scale features as input. Data augmentation is applied to both input images and BEV labels, following the approach described in PointBEV~\cite{chambon2024pointbev}. The BEV features rendering is implemented based on the original Gaussian Splatting library~\cite{kerbl3Dgaussians}. The error coefficient \( k \) is empirically set to 0.5, and the number of multi-scale BEV features is set to 3, corresponding to resolutions of \( 50 \times 50 \), \( 100 \times 100 \), and \( 200 \times 200 \). Unless otherwise specified, our experiments are conducted using an EfficientNet-B4 backbone~\cite{tan2020efficientnetrethinkingmodelscaling}, with an image resolution of \( 224 \times 480 \), and without applying visibility filtering for low-visibility vehicles.

\input{sec/table1}
\input{sec/table2}

\subsection{Comparisons with Existing Methods}

We compare GaussianLSS with both unprojection-based and projection-based approaches. As shown in Table \ref{table:1}, GaussianLSS outperforms all unprojection-based methods and achieves comparable performance to projection-based methods. We observe that even if the utilization of multi-scale rendering, GaussianLSS predicts distorted in terms of object shape poorer than projection-based methods, but could capture farther objects. We will show it in the next section. Moreover, Table \ref{tab:pedestrian_segmentation} compares the pedestrian class segmentation, while Table \ref{tab:inference} shows the inference speed and memory consumption. GaussianLSS achieves 80.2 FPS which is $2.5\times$ faster than PointBEV, showcasing its efficiency. Beyond the listed tasks, we also evaluate GaussianLSS on other applications, including map segmentation and 3D object detection, as presented in the supplementary material. These results further validate the versatility of GaussianLSS in different tasks, showing the effectiveness of uncertainty representation.

\input{sec/table3}
\input{sec/table4}
\input{sec/FIG4}

\subsection{Ablations with Error Tolerance}

The error tolerance coefficient \( k \) is a critical parameter in our uncertainty modeling approach. It defines the spatial extent of the 3D uncertainty representation by controlling the size of the ellipsoid surrounding each 3D point. A smaller \( k \) restricts the splatting to high-confidence regions but risks ignoring valid extent of the objects, while a larger \( k \) incorporates more uncertain areas at the expense of precision. This balance is essential for effective BEV feature representation. The analysis is shown in Figure~\ref{fig:error_tolerance}. On the other hand, we also experiment with directly predicting a fixed spatial extent instead of modeling the uncertainty. By comparing these two approaches, we aim to highlight the advantages of incorporating uncertainty into the BEV feature extraction process, as predicting extent yields 1.3\% lower performance.

\input{sec/table5}
\input{sec/table6}

\subsection{Depth Uncertainty Modeling Analysis}
We evaluate GaussianLSS's performance across varying distances from the ego-vehicle, focusing on its capability to accurately represent long-range objects. For this experiment, we calculate the IoU while excluding predictions within certain proximity thresholds to the ego-vehicle. Figure~\ref{fig:distance} compares GaussianLSS with the state-of-the-art projection-based approach, PointBEV~\cite{chambon2024pointbev}. Both models exhibit similar error trends, with decreasing accuracy as the distance increases due to growing depth ambiguity. However, GaussianLSS demonstrates an advantage in handling long-range objects. By explicitly modeling and leveraging depth uncertainty, GaussianLSS provides more accurate object representations beyond 30 meters. The performance drops to zero as the maximum depth is set to 61 meters.

\input{sec/FIG5}

\subsection{Opacity in Feature Rendering}
Opacity in feature rendering can be viewed as a weighted sum mechanism, guiding the model to focus on semantically relevant regions. High-opacity regions represent confident, high-contribution features, while low-opacity regions are de-emphasized or can be filtered out for efficiency. This adaptive mechanism allows the model to learn to prioritize meaningful areas, while reducing the influence of less important regions. Empirically, after training, 80\% of the Gaussians have an opacity below 0.01, highlighting the model's efficiency in identifying and projecting key regions into the BEV space, as illustrated in Figure~\ref{fig:opacity}.

\subsection{Qualitative Results}
We show qualitative results in Figure~\ref{fig:qualitative_results1}.  The yellow regions indicate areas masked out during feature lifting due to low opacity values, ensuring the model to focus on semantically significant features. GaussianLSS captures critical regions such as vehicles, even in challenging urban scenes with occlusions and clutters. This highlights the effectiveness of GaussianLSS in learning meaningful features while filtering irrelevant regions, leading to accurate and efficient BEV representations.
Figure~\ref{fig:qualitative_results2} presents 8 different scenarios of the model's robustness in long-range perception. We visualize the uncertainty-aware features after splatting onto the BEV plane. Despite the inherent challenges of long-range perception, including the increased depth ambiguity, GaussianLSS leverages uncertainty-aware features to focus on regions of interest while maintaining accuracy in BEV projection.

%% file: sec/table1.tex
\begin{table}[t]
    \centering
    \caption{\textbf{BEV segmentation IoU for Vehicle on the nuScenes dataset.} We compare GaussianLSS with multiple existing approaches across 4 different settings, incorporating visibility filtering and two resolution configurations, following PointBEV \cite{chambon2024pointbev}. The upper rows represent projection-based baselines, while the lower rows correspond to unprojection-based methods. GaussianLSS achieves state-of-the-art performance against unprojection-based baselines across all settings.}
    \resizebox{\linewidth}!{%
    \begin{tabular}{lcccccc}
        \hline
        \multicolumn{2}{c}{\textbf{Vehicle segm. IoU} ($\uparrow$)} & \multicolumn{2}{c}{\textbf{No visibility filtering}} & \multicolumn{2}{c}{\textbf{Visibility filtering}} \\
        \textbf{Method} & \textbf{Backbone.} & \textbf{224 $\times$ 480} & \textbf{448 $\times$ 800} & \textbf{224 $\times$ 480} & \textbf{448 $\times$ 800} \\
        \hline
        BEVFormer \cite{li2022bevformer} & RN-50 & 35.8 & 39.0 & 42.0 & 45.5 \\
        Simple-BEV \cite{harley2022simple} & RN-50 & 36.9 & 40.9 & 43.0 & 44.9 \\
        PointBeV \cite{chambon2024pointbev} & EN-b4 & \textbf{38.7} & \textbf{42.1} & \textbf{44.0} & \textbf{47.6} \\
        \hline
        FIERY static \cite{fiery2021} & EN-b4 & 35.8 & --- & 39.8 & --- \\
        CVT \cite{CVT} & EN-b4 & 31.4 & 32.5 & 36.0 & 37.7 \\
        LaRa \cite{bartoccioni2022lara} & EN-b4 & 35.4 & --- & 38.9 & --- \\
        BAEFormer \cite{Pan_2023_CVPR} & EN-b4 & 36.0 & 37.8 & 38.9 & 41.0 \\
        GaussianLSS & EN-b4 & \textbf{38.3} & \textbf{40.6} & \textbf{42.8} & \textbf{46.1} \\
        \hline
    \end{tabular}%
    }
\label{table:1}
\vspace{2mm}
\end{table}

%% file: sec/table2.tex
\begin{table}[t]
    \centering
    \small
    \caption{\textbf{BEV pedestrian segmentation on nuScenes.} GaussianLSS performs favorably against all unprojection-based baselines and is only 1.1\% behind the state-of-the-art model. The experiments are conducted at a resolution of $224\times480$ with visibility filtering.}
    \begin{tabular}{lcc}
        \toprule
        \textbf{Pedestrian segm.} & 
        \textbf{IoU (\(\uparrow\))} \\
        \midrule
        BEVFormer \cite{harley2022simple} & 16.4 \\
        SimpleBEV \cite{li2022bevformer} & 17.1 \\
        PointBeV \cite{chambon2024pointbev}    & \textbf{18.5} \\
        \hline
        LSS \cite{philion2020lift}                      & 15.0 \\
        FIERY static\cite{fiery2021}                          & 17.2 \\
        CVT \cite{CVT}                          &       14.2 \\
        ST-P3 \cite{hu2022stp3}                         & 14.5 \\
        GaussianLSS & \textbf{18.0} \\
        \bottomrule
    \end{tabular}
    \label{tab:pedestrian_segmentation}
    \vspace{2mm}
\end{table}

%% file: sec/table3.tex
\begin{table}[!t]
    \centering
    \small
    \caption{\textbf{Comparisons of inference speed, memory consumption, and vehicle segmentation IoU}. GaussianLSS achieves comparable IoU (-0.4\%) performance to PointBEV while being significantly faster, demonstrating over 2.5x the inference speed and 0.3x memory consumption.}
    \begin{tabular}{lcccc}
        \toprule
        \textbf{Method} & 
        \textbf{FPS} &
        \textbf{Mem GiB} &
        \textbf{IoU}
        \\
        \midrule
        BEVFormer \cite{li2022bevformer}& 34.7 & 0.47 & 35.8 \\
        SimpleBeV \cite{harley2022simple}& 37.1 & 3.31 & 36.9 \\
        PointBeV \cite{chambon2024pointbev}& 32.0 & 1.26 & 38.7\\
        \midrule

        FIERY static \cite{fiery2021}& 27.3 & 0.40 & 35.8 \\
        CVT \cite{CVT} & 107.6 & 0.35 & 31.4 \\
        GaussianLSS & 80.2 & 0.33 & 38.3\\
        \bottomrule
    \end{tabular}
    \label{tab:inference}
    \vspace{2mm}
\end{table}

%% file: sec/table4.tex
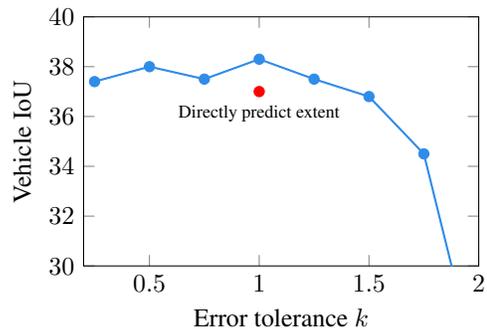
\begin{figure}[t]
    \centering
    \resizebox{0.8\linewidth}!{
    \begin{tikzpicture}
        \begin{axis}[
            width=7cm, height=5cm, 
            xlabel={Error tolerance $k$}, 
            ylabel={Vehicle IoU}, 
            xtick={0.5, 1.0, 1.5, 2.0},
            xmin=0.2, xmax=2.0, 
            ymin=30, ymax=40, 
            grid style={dashed, gray!30}, 
            legend style={font=\small}, 
        ]
        
        \addplot[color=bleudefrance, thick] coordinates {
            (0.25, 37.4)
            (0.5,  38.0)
            (0.75, 37.5)
            (1.0,  38.3)
            (1.25, 37.5)
            (1.5,  36.8)
            (1.75, 34.5)
            (2.0,  25.6)
        };
        \addplot[
        color=bleudefrance,
        mark=*,
        ]coordinates {
            (0.25, 37.4)
            (0.5,  38.0)
            (0.75, 37.5)
            (1.0,  38.3)
            (1.25, 37.5)
            (1.5,  36.8)
            (1.75, 34.5)
            (2.0,  25.6)
        };
        \addplot[
        color=red,
        mark=*,
        ]coordinates {
            (1.0,  37.0)
        };
        \node[below,font=\tiny] at (axis cs:1,36.8) {\scriptsize Directly predict extent};
        \end{axis}
    \end{tikzpicture}
    }
    \caption{\textbf{Sweeping analysis on error tolerance $k$.} We vary the error tolerance coefficient $k$ across a range of values ($k=[0.25,2.0]$). The results indicate that performance remains consistent for $k$ values between 0.5 and 1.25. However, when $k$ becomes too large, the IoU drops significantly as the model tolerates excessive ambiguity, causing the features to spread out too much and lose precision. The red dot represents the baseline approach of directly predicting the extent of the 3D mean.}
    \label{fig:error_tolerance}
\end{figure}

%% file: sec/FIG4.tex
\begin{figure*}[th]
    \centering

    \resizebox{0.95\linewidth}!{
    \begin{subfigure}[t]{\textwidth}
        \includegraphics[width=\textwidth]{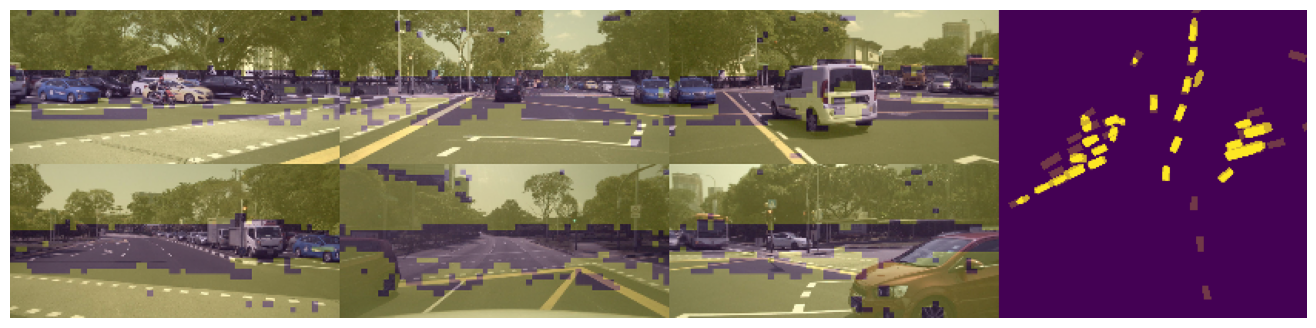}
    \end{subfigure}
    }

    \resizebox{0.95\linewidth}!{
    \begin{subfigure}[t]{\textwidth}
        \includegraphics[width=\textwidth]{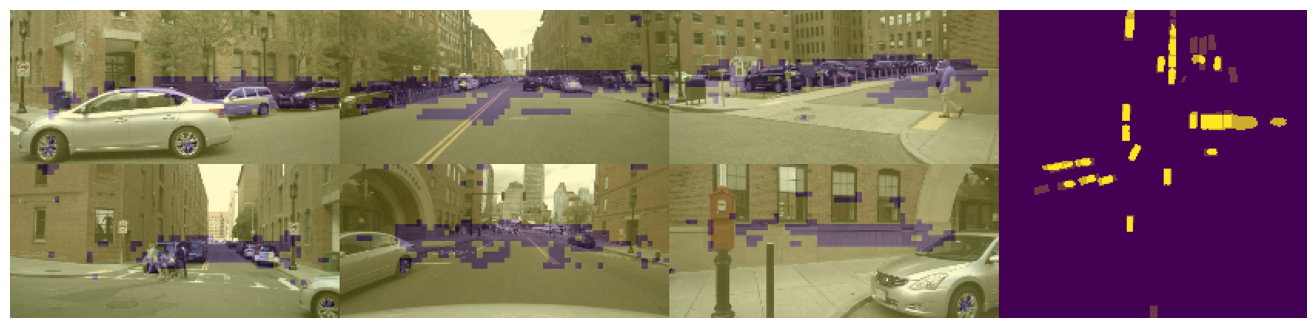}
    \end{subfigure}
    }
    \caption{\textbf{Qualitative results demonstrating the effectiveness of semantic learning by filtering opacity values below 0.01.} The yellow regions represent masked-out areas during features lifting. The left column shows the six camera views surrounding the ego-vehicle, with the top three views being front-facing and the bottom three being back-facing. The right column depicts BEV predictions overlapped with the ground truth segmentation for reference. The results demonstrate the model's ability to learn meaningful semantic features and accurately project relevant regions to the BEV plane. The ego-vehicle is centered in the map, with visualization highlights focusing on critical areas.}
    \label{fig:qualitative_results1}
    \vspace{2mm}
\end{figure*}

%% file: sec/table5.tex
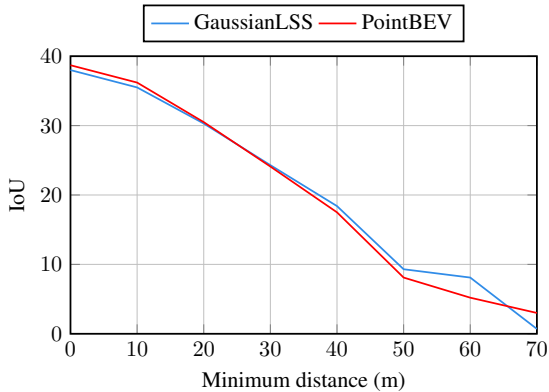
\begin{figure}[t]
\centering
\resizebox{0.9\linewidth}!{
\begin{tikzpicture}
\begin{axis}[
    width=9cm, height=6cm,
    xlabel={Minimum distance (m)},
    ylabel={IoU},
    xmin=0, xmax=70,
    ymin=0, ymax=40,
    xtick={0, 10, 20, 30, 40, 50, 60, 70},
    ytick={0, 10, 20, 30, 40},
    grid=major,
    legend style={at={(0.5,1.05)}, anchor=south, legend columns=2},
    legend cell align={left},
    thick
]

\addplot[
    color=bleudefrance,
    solid,
    thick
] coordinates {
    (0, 38.0) 
    (10, 35.5) 
    (20, 30.3)
    (30, 24.3)
    (40, 18.4)
    (50, 9.3) 
    (60, 8.1) 
    (70, 0.7)
};
\addlegendentry{GaussianLSS}

\addplot[
    color=red,
    solid,
    thick
] coordinates {
    (0, 38.7) 
    (10, 36.2) 
    (20, 30.5) 
    (30, 24.1) 
    (40, 17.5)
    (50, 8.1)
    (60, 5.2)
    (70, 3.0)
};
\addlegendentry{PointBEV}

\end{axis}
\end{tikzpicture}
}
\caption{\textbf{Impact of distance on vehicle segmentation IoU}. We compare between IoU and distance to the ego-vehicle. Each marker represents the average IoU for vehicles at least $d$ meters away.}
\label{fig:distance}
\end{figure}

%% file: sec/table6.tex
\begin{figure}[t]
\centering
\begin{tikzpicture}
\begin{axis}[
    width=8cm, height=5cm,
    xlabel={Epochs},
    ylabel={Retained Gaussians (\%)},
    xmin=0, xmax=50,
    ymin=0, ymax=100,
    xtick={0, 10, 20, 30, 40, 50},
    ytick={0, 20, 40, 60, 80, 100},
    grid=major,
    legend pos=north east,
    legend style={anchor=west, at={(0.382,0.5)}}
]

\addplot[
    color=bleudefrance,
    solid,
    thick
] coordinates {
    (0, 92.3) 
    (5, 32.2) 
    (10, 25.7)
    (15, 21.7)
    (20, 19.0)
    (25, 17.8) 
    (30, 16.7) 
    (35, 15.3)
    (40, 14.0)
    (45, 13.3)
    (50, 13.2)
};

\addlegendentry{Retained Gaussians}
\end{axis}

\begin{axis}[
    width=8cm, height=5cm,
    xmin=0, xmax=50,
    ymin=30, ymax=40,
    xlabel={Epochs},
    ylabel={Vehicle IoU (\%)},
    axis y line*=right, 
    ytick={30, 32, 34, 36, 38, 40},
    legend pos=north east,
    legend style={anchor=west, at={(0.55,0.3)}}
]

\addplot[
    color=red,
    solid,
    thick
] coordinates {
    (0, 28.6)
    (5, 32.5)
    (10, 34.8)
    (15, 35.6)
    (20, 36.2)
    (25, 37.0)
    (30, 37.3)
    (35, 37.5)
    (40, 37.6)
    (45, 37.8)
    (50, 38.3)
};
\addlegendentry{Vehicle IoU}

\end{axis}

\end{tikzpicture}
\caption{\textbf{Proportion of Gaussians retained and Vehicle IoU over training epochs.} The proportion of retained Gaussians($\alpha<0.01$) reduces to around 20\% as the model converges, improving efficiency. Meanwhile, Vehicle IoU steadily increases, showcasing the improved semantic accuracy.}
\label{fig:opacity}
\vspace{2mm}
\end{figure}
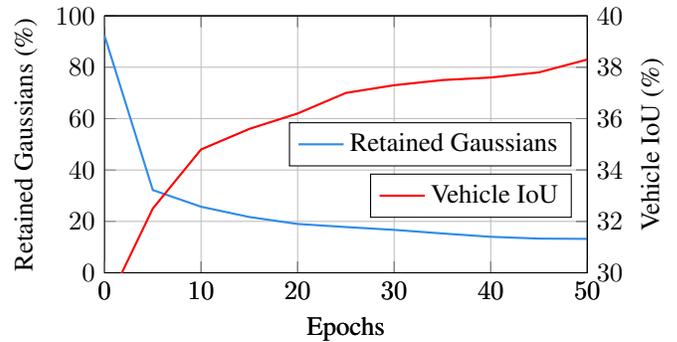

%% file: sec/FIG5.tex
\begin{figure*}[th]
    \centering
    

    \begin{subfigure}[t]{0.475\textwidth}
        \centering
        \includegraphics[width=\linewidth]{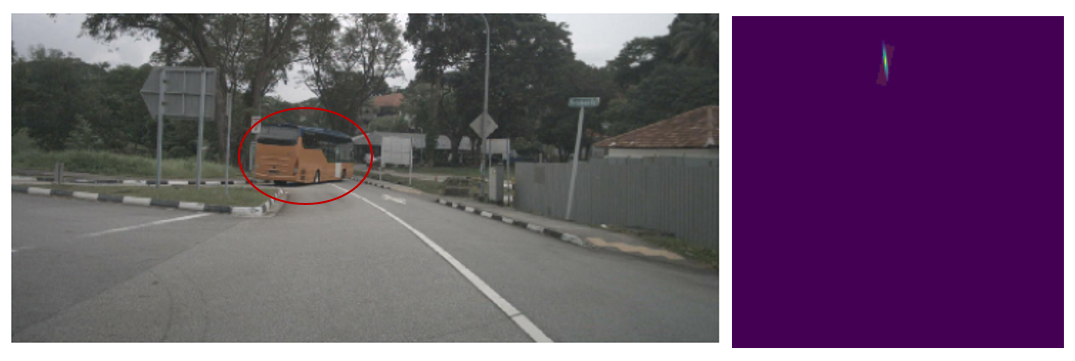}
    \end{subfigure}
    \hspace{0.1cm}
    \begin{subfigure}[t]{0.475\textwidth}
        \centering
        \includegraphics[width=\linewidth]{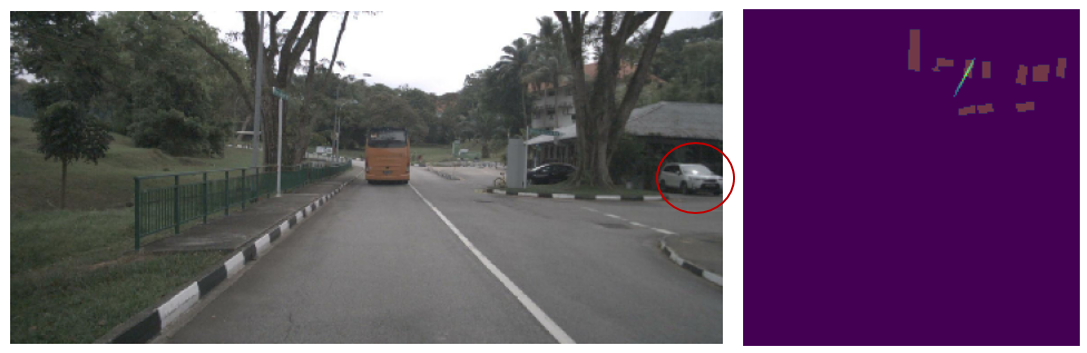}
    \end{subfigure}

    \vspace{-0.5cm} 

    \begin{subfigure}[t]{0.475\textwidth}
        \centering
        \includegraphics[width=\linewidth]{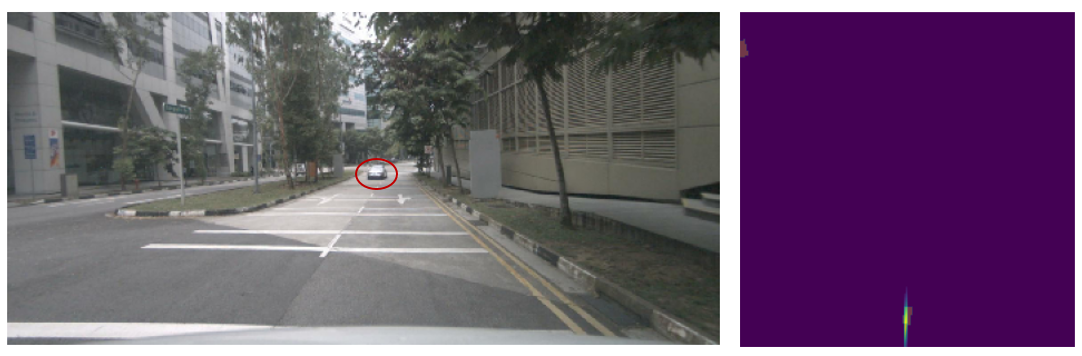}
    \end{subfigure}
    \hspace{0.1cm}
    \begin{subfigure}[t]{0.475\textwidth}
        \centering
        \includegraphics[width=\linewidth]{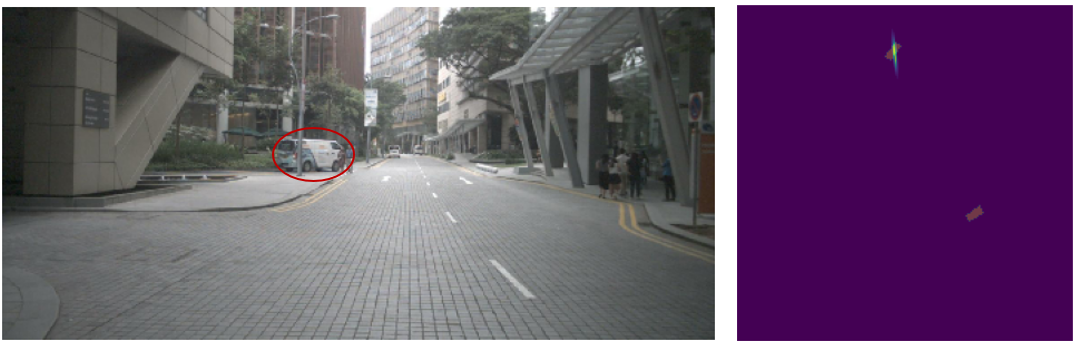}
    \end{subfigure}

    \vspace{-0.5cm} 

    \begin{subfigure}[t]{0.475\textwidth}
        \centering
        \includegraphics[width=\linewidth]{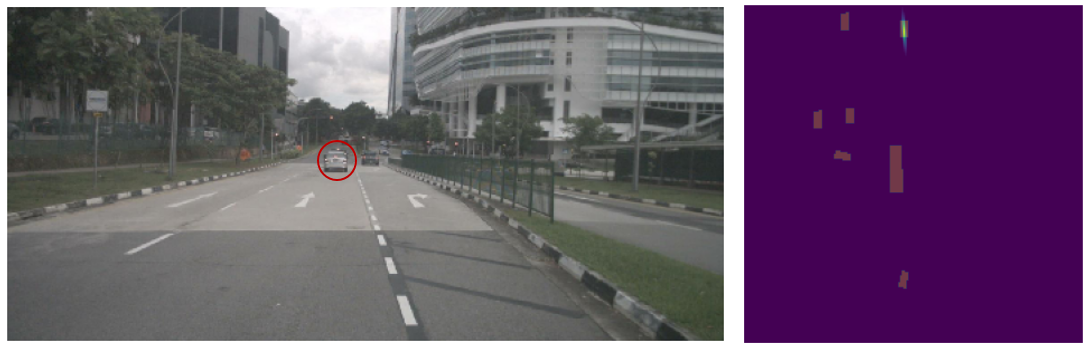}
    \end{subfigure}
    \hspace{0.1cm}
    \begin{subfigure}[t]{0.475\textwidth}
        \centering
        \includegraphics[width=\linewidth]{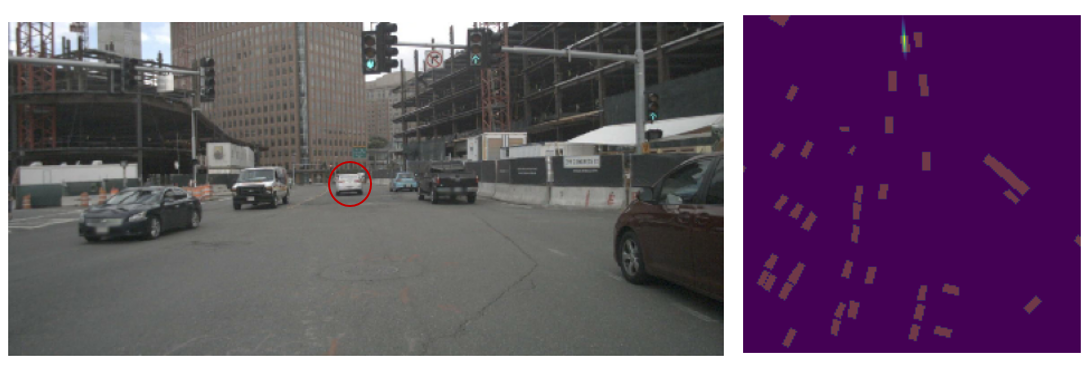}
    \end{subfigure}
    
    \vspace{-0.5cm}
    
    \begin{subfigure}[t]{0.475\textwidth}
        \centering
        \includegraphics[width=\linewidth]{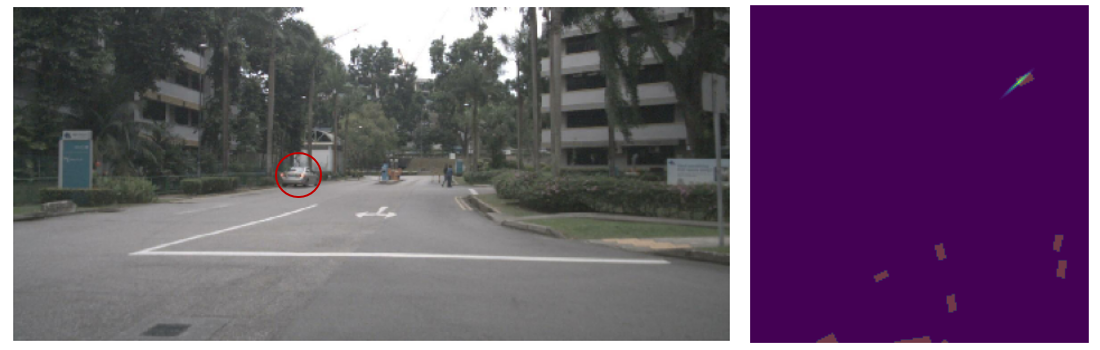}
    \end{subfigure}
    \hspace{0.1cm}
    \begin{subfigure}[t]{0.475\textwidth}
        \centering
        \includegraphics[width=\linewidth]{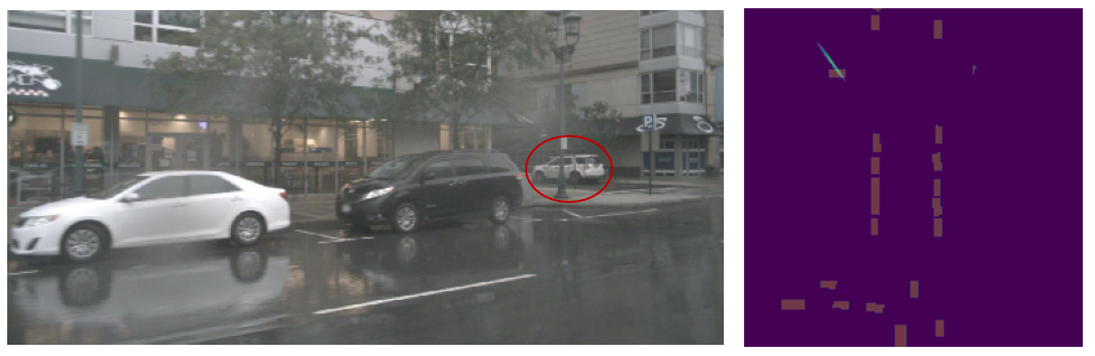}
    \end{subfigure}
    
    \caption{\textbf{Visualization of our uncertainty-aware feature splatting approach.} The circled vehicles are located at distances greater than 30 meters from the ego-vehicle. The right column illustrates how the circled areas are projected into the BEV space. Despite the inherent challenges of long-range perception, our model effectively identifies the vehicles' positions without relying on precise depth estimation, demonstrating its robustness in managing depth uncertainty.}
    \label{fig:qualitative_results2}
    \vspace{1mm}
\end{figure*}

%% file: sec/5_conclusion.tex
\vspace{3mm}
\section{Conclusions}

We presented GaussianLSS, a novel approach to BEV perception that integrates depth uncertainty modeling with efficient multi-scale BEV feature rendering. 
By transforming per-pixel depth uncertainty into 3D Gaussian representations, GaussianLSS effectively addresses the inherent challenges of depth ambiguity while enabling robust and accurate feature projection into the BEV space.
We achieve state-of-the-art performance among unprojection-based methods with significant memory efficiency and inference speed. These results validate the potential of depth uncertainty modeling in enhancing BEV perception for real-world autonomous driving applications. Future work will explore its applicability to other BEV-based tasks and temporal setting.

%% file: X_suppl.tex
\clearpage
\setcounter{page}{1}
\maketitlesupplementary
\appendix






\section{BEV Feature with Gaussian Splatting}
\label{sec:BEVSplat}

\subsection{Projection}

As described in Gaussian Splatting~\cite{kerbl3Dgaussians}, to render 3D Gaussians in image space, we project their covariance matrix $\Sigma$ using a viewing transformation $W$ and the Jacobian $J$ of the affine approximation of the projective transformation:

\[
\Sigma' = J W \Sigma W^T J^T,
\]
where $\Sigma'$ is the projected covariance matrix.
Projecting to the Bird's Eye View (BEV) image space simplifies the process significantly because the $z$-axis can be ignored. The BEV scaling matrix, $S_{\text{BEV}}$, scales the 3D coordinates to the 2D BEV plane and is defined as:

\[
S_{\text{BEV}} =
\begin{bmatrix}
0 & \text{scale}_x \\
\text{scale}_y & 0 \\
\end{bmatrix}.
\] 
Note that the \( x \) and \( y \) axes of the 3D coordinates are swapped when mapping to the BEV plane.
The projection of the covariance matrix $\Sigma$ into the BEV image space using the scaling matrix $S_{\text{BEV}}$ is then given by:

\[
\Sigma' = S_{\text{BEV}} \Sigma_{xy} S_{\text{BEV}}^T,
\]
where $\Sigma_{xy}$ is the $2 \times 2$ submatrix of $\Sigma$ corresponding to the $x$ and $y$ axes. Therefore, the projected covariance matrix $\Sigma'$ becomes:

\[
\Sigma' =
\begin{bmatrix}
\Sigma_{22} \cdot \text{scale}_x^2 & \Sigma_{21} \cdot \text{scale}_x \cdot \text{scale}_y \\
\Sigma_{12} \cdot \text{scale}_x \cdot \text{scale}_y & \Sigma_{11} \cdot \text{scale}_y^2 \\
\end{bmatrix}.
\]

\subsection{Gradient Computation}

Next, we compute the derivative of the loss $L$ with respect to the covariance matrix $\Sigma$, denoted as $\frac{\partial L}{\partial \Sigma}$. Let $\Sigma'$ be defined as:

\[
\Sigma' =
\begin{bmatrix}
a & b \\
b & c
\end{bmatrix},
\]
where:

\[
\begin{aligned}
a &= \Sigma_{22} \cdot \text{scale}_x^2, \\
b &= \Sigma_{21} \cdot \text{scale}_x \cdot \text{scale}_y, \\
c &= \Sigma_{11} \cdot \text{scale}_y^2.
\end{aligned}
\]

Using the chain rule, the gradient $\frac{\partial L}{\partial \Sigma_{ij}}$ is given by:

\[
\frac{\partial L}{\partial \Sigma_{ij}} = \frac{\partial L}{\partial a} \frac{\partial a}{\partial \Sigma_{ij}} + \frac{\partial L}{\partial b} \frac{\partial b}{\partial \Sigma_{ij}} + \frac{\partial L}{\partial c} \frac{\partial c}{\partial \Sigma_{ij}}.
\]

We compute the partial derivatives:

- For $\Sigma_{11}$:

  \[
  \frac{\partial a}{\partial \Sigma_{11}} = 0, \quad \frac{\partial b}{\partial \Sigma_{11}} = 0, \quad \frac{\partial c}{\partial \Sigma_{11}} = \text{scale}_y^2.
  \]

  Therefore:

  \[
  \frac{\partial L}{\partial \Sigma_{11}} = \frac{\partial L}{\partial c} \cdot \text{scale}_y^2.
  \]

- For $\Sigma_{12}$ (since $\Sigma$ is symmetric, $\Sigma_{12} = \Sigma_{21}$):

  \[
  \frac{\partial a}{\partial \Sigma_{12}} = 0, \quad \frac{\partial b}{\partial \Sigma_{12}} = \text{scale}_x \cdot \text{scale}_y, \quad \frac{\partial c}{\partial \Sigma_{12}} = 0.
  \]

  Therefore:

  \[
  \frac{\partial L}{\partial \Sigma_{12}} = \frac{\partial L}{\partial b} \cdot \text{scale}_x \cdot \text{scale}_y.
  \]

- For $\Sigma_{22}$:

  \[
  \frac{\partial a}{\partial \Sigma_{22}} = \text{scale}_x^2, \quad \frac{\partial b}{\partial \Sigma_{22}} = 0, \quad \frac{\partial c}{\partial \Sigma_{22}} = 0.
  \]

  Therefore:

  \[
  \frac{\partial L}{\partial \Sigma_{22}} = \frac{\partial L}{\partial a} \cdot \text{scale}_x^2.
  \]
Hence, the gradient $\frac{\partial L}{\partial \Sigma}$ is:

\[
\frac{\partial L}{\partial \Sigma} =
\begin{bmatrix}
\displaystyle \frac{\partial L}{\partial c} \cdot \text{scale}_y^2 & \displaystyle \frac{\partial L}{\partial b} \cdot \text{scale}_x \cdot \text{scale}_y \\
\displaystyle \frac{\partial L}{\partial b} \cdot \text{scale}_x \cdot \text{scale}_y & \displaystyle \frac{\partial L}{\partial a} \cdot \text{scale}_x^2
\end{bmatrix}.
\]
\section{Ablation Study on Depth Prediction}
We conduct an ablation study to evaluate the impact of various depth settings, including depth ranges and bin sizes, on model performance, as shown in Table~\ref{tab:depthablation}. The result demonstrates that the model’s performance is relatively stable across different depth range and bin size configurations, with the setting (1, 61) and bin size 64 providing a slightly higher IoU. This robustness simplifies parameter selection in practice.
\input{sec/table_supp_1}

\section{Map Segmentation and 3D Object Detection}

Beyond instance-level BEV segmentation, we further evaluate GaussianLSS on \textbf{map segmentation} and \textbf{3D object detection} to assess its generalization capability. For map segmentation, we predict drivable areas, pedestrian crossings, walkways, and road dividers, following the same experimental setup as prior works. As shown in Table \ref{table:map}, GaussianLSS achieves competitive performance.

For 3D object detection, we integrate a detection head directly into the BEV features, following the approach of BEVFormer. We evaluate performance using mean Average Precision (mAP) and nuScenes Detection Score (NDS). As shown in Table \ref{table:det3d}, our method extends beyond BEV segmentation and is also applicable to detection tasks. 

\input{rebuttal/map_segmentation}
\input{rebuttal/Det3D}
\section{Submodule Speed Analysis}

To further evaluate the efficiency of our approach, we analyze the runtime performance of key submodules and compare them with baseline methods. We break down the inference time into different processing stages, including the backbone, neck, view transformation, and head. Table \ref{table:module_speed} reports the speed of each submodule.
Our method demonstrates a significant speed advantage in the view transformation stage while maintaining comparable efficiency to projection-based methods.
\input{rebuttal/component_speed}

\section{Model Structure}

We adopt three parallel CNN branches to predict features, opacity, and depth distributions, respectively. Each branch is composed of three convolution blocks (Conv-BN-ReLU). The BEV features are subsequently processed by a lightweight U-Net-like BEV decoder before feeding it into the task-specific head. Our decoder design is the same as prior works\cite{chambon2024pointbev, philion2020lift, harley2022simple} with lower parameters.

%% file: sec/table_supp_1.tex
\begin{table}[t]
    \centering
    \caption{\textbf{Ablation study on depth settings.} The first depth range, \((1, 61)\), where the minimum depth is set to 1 meter and the maximum depth is set to 61 meters, corresponds to settings used in the paper. The second depth range, \((0.5, 71)\), represents an extended range capturing the minimum and maximum depth values in the BEV plane. Bin sizes \((32, 64, 128)\) are also evaluated for their impact on IoU for the vehicle class.}
     \begin{tabular}{lcc}
        \toprule
        \textbf{Depth range} & 
        \textbf{Bin size $B$} &
        \textbf{IoU Vehicle}\\
        \midrule
        $(1, 61)$ & $32$ & 37.7              \\
        $(1, 61)$ & $64$ & \textbf{38.3}     \\
        $(1, 61)$ & $128$ & 37.8             \\

        \midrule
        $(0.5, 71)$ & $32$ & 37.7   \\
        $(0.5, 71)$ & $64$ & 37.9   \\
        $(0.5, 71)$ & $128$ & 37.6  \\

        \bottomrule
    \end{tabular}
    \label{tab:depthablation}
    \vspace{2mm}
\end{table}

%% file: rebuttal/map_segmentation.tex
\begin{table}[!h]
    \centering
    \caption{\textbf{Map Segmentation Comparisons.} We evaluate our method for common map classes on nuScenes.}
    \renewcommand{\arraystretch}{1.2} 
    \setlength{\tabcolsep}{3pt} 
    \vspace{-3mm}
    \resizebox{0.8\linewidth}{!}{ 
    \begin{tabular}{lcccc} 
        \hline
        \small \textbf{Method} & \small \textbf{Drivable} & \small \textbf{Ped. Cross.} & \small \textbf{Walkway} & \small \textbf{Divider} \\
        \hline
        LSS \cite{philion2020lift}  & 75.4 & 38.8 & 46.3 & 36.5 \\
        CVT \cite{CVT}  & 74.3 & 36.8 & 39.9 & 29.4 \\
        GaussianLSS & \textbf{76.3} & \textbf{46.3} & \textbf{50.2} & \textbf{38.7} \\
        \hline
    \end{tabular}
    }
    \label{table:map}
\end{table}

%% file: rebuttal/Det3D.tex
\begin{table}[!h]
    \centering
    \caption{\textbf{3D Detection Performance on nuScenes Validation.} 
    \vspace{-2mm}
    }
    {%
    \begin{tabular}{lcc}
        \hline
        \textbf{Method} & NDS & mAP \\
        \hline
        BEVDet \cite{huang2021bevdet}
        & 35.0 & 28.3   \\
        BEVFormer \cite{li2022bevformer}
        & 35.4 & 25.2   \\
        GaussianLSS 
        & 34.0 & 26.6   \\
        \hline
    \end{tabular}}
    \vspace{-2mm}
\label{table:det3d}
\vspace{1mm}
\end{table}

%% file: rebuttal/component_speed.tex
\begin{table}[!h]
    \centering
    \caption{\textbf{Comparison of Submodule Execution Time.} All times are measured in milliseconds. The ``VT'' column represents the view-transformation module (BEV encoder). All measurements are conducted on an RTX 4090 GPU.}
    \renewcommand{\arraystretch}{1.2} 
    \setlength{\tabcolsep}{4pt} 
    \small 
    \resizebox{1.0\linewidth}{!}{
    \begin{tabular}{lccccc} 
        \hline
        \textbf{Method} & \textbf{Backbone} & \textbf{Neck} & \textbf{VT} & \textbf{Head} & \textbf{Total} \\
        \hline
        PointBEV \cite{chambon2024pointbev}       & 5.6  & 0.21 & 13.47 & 1.95 & 21.23 \\
        FIERY static \cite{fiery2021}   & 5.71 & \textemdash{} & 36.46 & 1.18 & 43.35 \\
        CVT \cite{CVT}           & 5.68 & \textemdash{} & 2.17  & 0.71 & 8.02 \\
        GaussianLSS              
        & 5.86 & 1.06 & 3.06  & 1.24 & 11.22 \\
        \hline
    \end{tabular}
    }
    \label{table:module_speed}
\end{table}